%% file: main.tex
\documentclass[10pt, conference]{IEEEtran}
\IEEEoverridecommandlockouts
\usepackage[dvipsnames,table,x11names]{xcolor}
\usepackage{cite}
\usepackage{amsmath,amssymb,amsfonts}
\usepackage{algorithmic}
\usepackage{graphicx,dblfloatfix}
\usepackage{textcomp}
\usepackage{xcolor}
\usepackage{multirow}
\usepackage{booktabs}
\usepackage[hidelinks]{hyperref}

\def\BibTeX{{\rm B\kern-.05em{\sc i\kern-.025em b}\kern-.08em
    T\kern-.1667em\lower.7ex\hbox{E}\kern-.125emX}}
\begin{document}

\title{Action Recognition in Video Recordings from Gynecologic Laparoscopy\thanks{This work was funded by the FWF Austrian Science Fund under grant P~32010-N38.}}

\author{\IEEEauthorblockN{Sahar Nasirihaghighi}
\IEEEauthorblockA{\textit{Institute of Information Technology (ITEC)} \\
\textit{Klagenfurt University, Austria}\\
Sahar.Nasirihaghighi@aau.at}
\and
\IEEEauthorblockN{Negin Ghamsarian}
\IEEEauthorblockA{\textit{Center for AI in Medicine} \\
\textit{University of Bern, Switzerland}\\
negin.ghamsarian@unibe.ch}
\and
\IEEEauthorblockN{Daniela Stefanics}
\IEEEauthorblockA{\textit{Institute of Information Technology (ITEC)} \\
\textit{Klagenfurt University, Austria}\\ 
Daniela.Stefanics@aau.at}
\and
\IEEEauthorblockN{Klaus Schoeffmann}
\IEEEauthorblockA{\textit{dInstitute of Information Technology (ITEC)} \\
\textit{Klagenfurt University, Austria}\\ 
Klaus.Schoeffmann@aau.at}
\and
\IEEEauthorblockN{Heinrich Husslein}
\IEEEauthorblockA{\textit{Department of Gynecology and Gynecological Oncology} \\
\textit{Medical University Vienna, Austria}\\
heinrich.husslein@meduniwien.ac.at}
}

\maketitle

\begin{abstract}
Action recognition is a prerequisite for many applications in laparoscopic video analysis including but not limited to 
surgical training, operation room planning, follow-up surgery preparation, post-operative surgical assessment, and surgical outcome estimation. However, automatic action recognition in laparoscopic surgeries involves numerous challenges such as (I) cross-action and intra-action duration variation, (II) relevant content distortion due to smoke, blood accumulation, fast camera motions, organ movements, object occlusion, and (III) surgical scene variations due to different illuminations and viewpoints. Besides, action annotations in laparoscopy surgeries are limited and expensive due to requiring expert knowledge. 
In this study, we design and evaluate a CNN-RNN architecture as well as a customized training-inference framework to deal with the mentioned challenges in laparoscopic surgery action recognition. Using stacked recurrent layers, our proposed network takes advantage of inter-frame dependencies to negate the negative effect of content distortion and variation in action recognition. Furthermore, our proposed frame sampling strategy effectively manages the duration variations in surgical actions to enable action recognition with high temporal resolution. Our extensive experiments confirm the superiority of our proposed method in action recognition compared to static CNNs.

\end{abstract}


\input{01_introduction}

\input{02_related_work}

\input{03_methodology}

\input{04_experimental_settings}

\input{05_experimental_results}

\input{06_conclusion}

\bibliographystyle{unsrt}
\bibliography{ref}
\vspace{12pt}
\color{red}

\end{document}

%% file: 01_introduction.tex
\section{Introduction}
\label{sec: introduction}

Laparoscopy, also known as keyhole surgery, is a minimally invasive surgery that allows a surgeon to access the inside of the abdomen without having to make large incisions in the skin. Avoiding large incisions is possible thanks to the laparoscope, a small and flexible tube equipped with a light source and a camera, which relays images of the inside of the abdomen to a monitor~\cite{kumar2016minimally}. The images on the display are used as “the eye of the surgeon” and allow the operating surgeons to control the intervention and supervise actions performed with instruments~\cite{lux2010novel}. 
As opposed to traditional open surgeries, which involve larger incisions, laparoscopic surgery has several advantages, including less pain and discomfort after the surgery, less scarring and a faster recovery time, and lower risk of complications such as infections and bleeding~\cite{velanovich2000laparoscopic,fuentes2014complications}. The recorded videos of laparoscopic surgeries are used for purposes such as surgical skill and quality assessment (SQA), novice surgeon training, postoperative analysis of the surgery, and research with the aim of ultimately improving the quality and safety of these surgeries. In the case of real-time surgical video analysis, these results can also be used for follow-up-surgery preparation during the operation~\cite{schoeffmann2015keyframe}.

\begin{figure*}[t!]
    \centering
    \includegraphics[width=13cm]{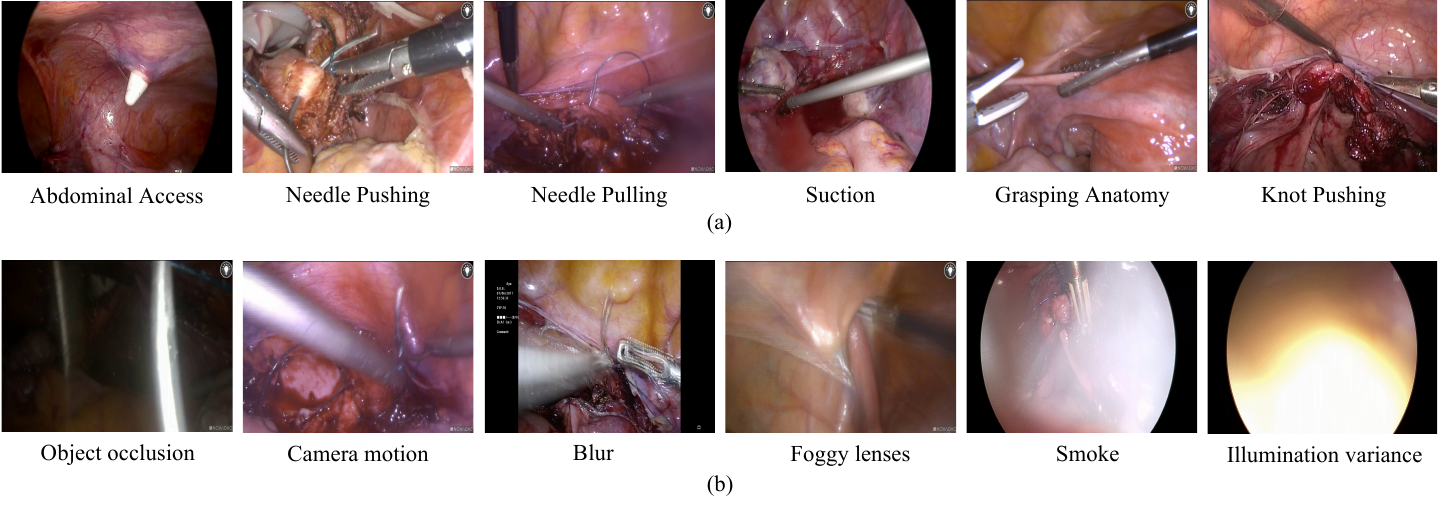}
    \caption{Sample frames of laparoscopic surgery videos. (a) different actions including abdominal access, needle pushing, needle pulling, suction, grasping anatomy, and knot pushing. (b) some challenges in laparoscopic surgery videos}
    \label{fig:1}
\end{figure*}

In recent years, a wide range of studies has been devoted to laparoscopic video analysis. These studies can be categorized into (I) surgical instrument segmentation and tracking ~\cite{lee2020evaluation}, and tool presence detection ~\cite{kondo2021lapformer}, (II) organ detection ~\cite{moccia2018uncertainty} and segmentation~\cite{maqbool2020m2caiseg}, (III) surgical phase recognition~\cite{kitaguchi2020real}, and (IV) surgical action recognition~\cite{bawa2021saras}. 

Surgical action recognition is a fundamental but challenging task in laparoscopic video analysis. It involves automatically identifying and classifying the various actions performed by the surgeon during the surgery. Depending on the surgical procedure, different surgical actions can happen in laparoscopic videos, such as abdominal access, coagulation, cutting, grasping anatomy, irritation, knot pushing, needle pulling, needle pushing, suction, and suturing. Fig.~\ref{fig:1}-(a) illustrates different actions in a laparoscopic surgery video.

Surgical action recognition in laparoscopic surgeries involves many challenges: (1) Different actions in a laparoscopic surgery follow a diverse distribution in terms of average duration, which leads to a highly imbalanced training set. (2) The duration of each independent action can have a wide range depending on the surgeon's skill level, the patient's conditions, the type of surgery, and so forth. In some cases, the relevant actions are too short, and detecting such actions is quite difficult. (3) Existence of smoke in the surgical region, foggy or bloody lenses, object occlusion, camera motion, respiratory motions of tissues, illumination variances, and viewpoint dependence~\cite{wang2018smoke} are the other hindering factors in laparoscopic action recognition. (4) Providing annotations for action recognition in these surgeries entails expert knowledge, and hence it is expensive.   Fig.~\ref{fig:1}-(b) illustrates some challenges in laparoscopic action recognition. The existence of such factors complicates the extraction of  underlying patterns from laparoscopic videos. Accordingly, designing an action recognition model that can deal with different challenges in laparoscopic videos requires extensive research. 

In this paper, we develop a robust model to recognize several relevant actions in laparoscopy, including \textit{abdominal access}, \textit{grasping anatomy}, \textit{knot pushing}, \textit{needle pulling}, and \textit{needle pushing}. The main contributions of this paper are: 

\begin{itemize}
    \item We present a comprehensive comparison of various neural network architectures' performance, including static CNNs, and end-to-end CNN-RNNs.
    \item We propose a CNN-RNN-based architecture to effectively deal with different challenges in detecting the target actions in laparoscopic surgery videos.
    \item Furthermore, we publicly release the dataset with our customized annotations to enable reproducing our results and support further investigations\footnote{\url{https://ftp.itec.aau.at/datasets/LapGyn6-Actions/}}.
\end{itemize}

The rest of the paper is organized as follows: in Section~\ref{sec: related work} the literature is reviewed for related work and differences to the current work are outlined. The proposed approach in this study is elaborated in Section~\ref{sec: methodology}. We explain the experimental setting in Section~\ref{sec: experimental settings} and report the experimental results in Section~\ref{sec: experimental results}. We conclude the paper in Section~\ref{sec: conclusion}.

%% file: 02_related_work.tex
\begin{figure*}[!t]
    \centering
    \includegraphics[width=0.8\textwidth]{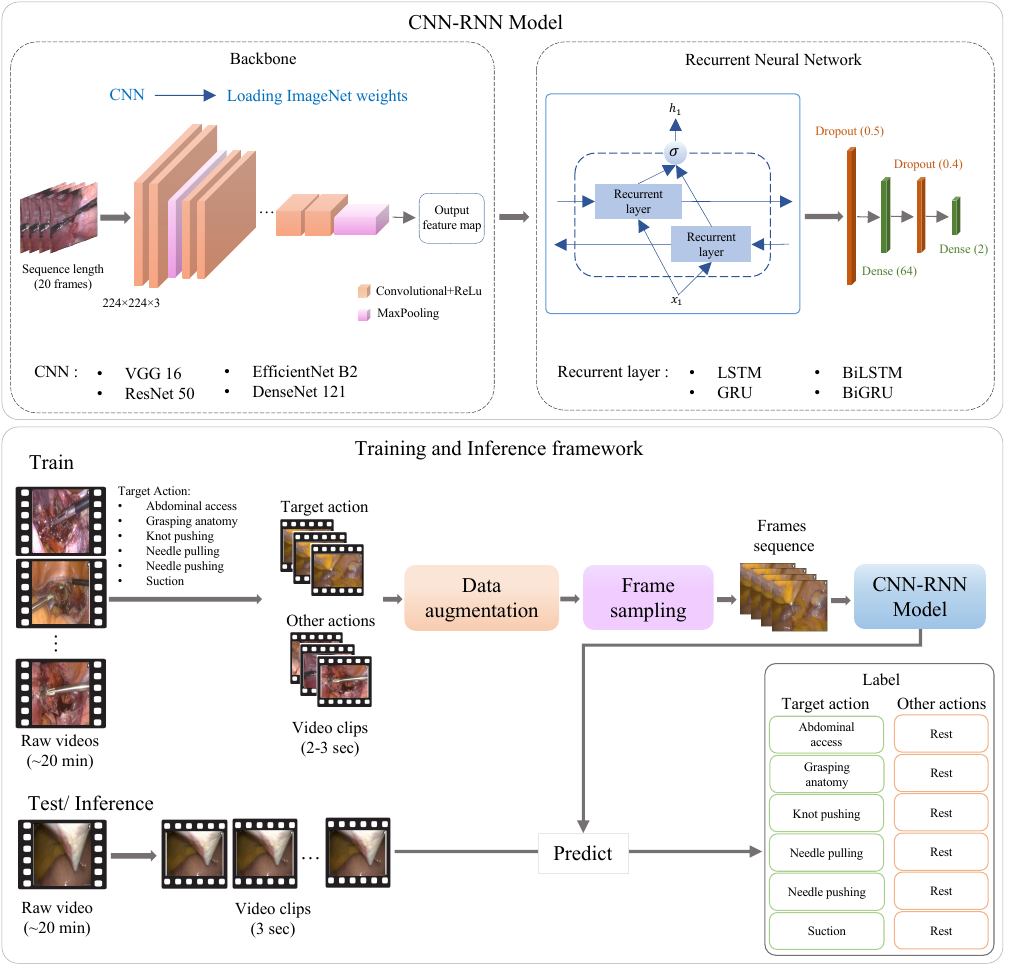}
    \caption{The Framework of the proposed model}
    \label{fig:Overview}
\end{figure*}

\section{Related Work}
\label{sec: related work}

This section summarizes state-of-the-art methods for laparoscopic surgical video analysis and action recognition in laparoscopy videos.

\paragraph{Laparoscopic Surgery Video Analysis}Since the current study involves CNN-RNNs, we summarize the methods that use recurrent layers for laparoscopy video analysis.
The study in ~\cite{namazi2022contextual} introduces LapTool-Net, a multi-label classification system, to detect surgical instruments in laparoscopic videos. The model is composed of a CNN (InceptionV3) as the spatial feature extractor and an RNN (GRU) to exploit the correlation between video frames. Jin et al.~\cite{jin2020multi} present a method for surgical phase recognition and tool presence detection. A deep CNN (Resnet) is used for feature extraction and tool presence detection, and an LSTM is applied for phase recognition. In ~\cite{namazi2018automatic}, SPD-DLS consisting of a CNN (InceptionV3) and an LSTM is proposed to recognize the surgical phases in laparoscopic videos. In ~\cite{golany2022artificial} Golany et al. propose a multi-stage temporal convolution network (MS-TCN) to incorporate temporal information of frames for surgical phase recognition. 
Sharghi et al. ~\cite{sharghi2022activity} present a surgical video activity detection method using a spatio-temporal model. A combination of a swin transformer as backbone with a GRU model is applied to recognize different phases.

\paragraph{Action Recognition in Laparoscopy Videos} Bawa et al.~\cite{bawa2021saras} present a feature pyramidal network model applying a residual network and a convolutional layer to predict the class score and the relevant bounding boxes. In ~\cite{huang2022surgical}, an online surgical action prediction and recognition system, named SA, is developed to help decision-making in the operating room.

In~\cite{petscharnig2018early}, the effect of early and late fusion of temporal information is investigated on surgical action recognition. Block-based motion estimation (BBME) and flow-based residual motion estimation (ResM) approaches are presented for early fusion. For the late fusion approach, the individual-frame predictions are accumulated.  In ~\cite{khatibi2020proposing}, three different models for automatically recognizing surgical action in laparoscopic videos are proposed. The models are a combination of pre-trained convolutional neural networks and machine learning algorithms for classifying the single and multi frames. 

\paragraph{Shortcomings of the existing approaches}Although existing approaches demonstrate good performance in surgical action recognition, they suffer from several drawbacks:
(1) To the best of our knowledge, there is no study on temporal action recognition with neural networks in gynecologic laparoscopy. The evaluations in \cite{bawa2021saras} and \cite{huang2022surgical} use traditional frame-based neural networks to recognize actions in radical prostatectomy. Similarly, in our previous work \cite{petscharnig2018early,petscharnig2018learning} we use frame-based models and optical flow estimation to recognize rather general laparoscopic actions and anatomy (e.g., \textit{Suture} and \textit{Ovary}). In \cite{khatibi2020proposing}, the authors utilize our publicly released dataset \textit{LapGyn4} \cite{leibetseder2018lapgyn4} to recognize laparoscopic actions with transformers, but this dataset is composed of frame samples (excerpts) from laparoscopic videos instead of full surgery videos, hence the results do not reflect the true achievable performance when the model is applied to every frame of a surgery recording. More concretely, the current methods cannot effectively utilize the temporal dependencies between close frames, which are imperative to action recognition. In this work, however, we focus on full video-based recognition of specific short-termed actions, such as \textit{Abdominal Access}, \textit{Needle Pulling}, and \textit{Knot Pushing}, etc., which are learned by incorporating spatiotemporal dependencies via a combination of convolutional and recurrent neural networks.     
(2) Imbalance is a common problem in surgical action recognition where some actions occur rarely and with a short duration, while other actions occur frequently. In case dataset imbalance is not addressed properly, the probability of the network's confirmation bias and overfitting to the majority classes will be high. To address this problem, we propose a novel training-and-inference framework based on data augmentation techniques and random sampling to balance the dataset and provide the model with diverse input data to boost the network's performance in action recognition. 
(3) In previous studies on laparoscopic action recognition, only the relevant actions were considered while ignoring the presence of irrelevant actions in the videos. However, in this study, we take into account both relevant and irrelevant actions present in the videos and classify the irrelevant action into the ``\textit{Rest}'' category alongside the other actions. This approach ensures more robustness of the trained network in discriminating the target actions from irrelevant actions.

%% file: 03_methodology.tex
\section{Methodology}
\label{sec: methodology}

In this section, we delineate our proposed CNN-RNN architecture for action recognition in laparoscopic surgery videos. We then explain our customized training and inference framework including dataset generation and input sequence sampling.

\subsection{CNN-RNN Model}
Fig.~\ref{fig:Overview} depicts the overview of the proposed framework, which consists of two main components: a convolutional neural network (CNN) as the backbone, and a recurrent neural network (RNN). 
The backbone model is responsible for extracting discriminative features from independent frames within the video. 
For this purpose, four different pre-trained CNN architectures, namely VGG16, ResNet50, EfficientNetB2, and DenseNet121, are utilized and compared. The input clip of the CNN network corresponds to one of the following two classes: (1) the video segments of the target action, and (2) video segments of the rest actions including relevant and irrelevant actions in the video. 

The recurrent neural network (RNN) component of our CNN-RNN model accounts for extracting temporal features from the input video segment. The input to the RNN layer is the output feature map of the backbone network. Indeed, the output of the last convolutional layer in the pre-trained CNN undergoes an average-pooling layer before being fed into the recurrent layer, which then produces a sequence of hidden states capturing the temporal dynamics of the video frames.
As shown in Fig.~\ref{fig:Overview}, we stacked two recurrent layers allowing the network to learn complex temporal patterns from the sample frames in each input sequence. By stacking layers, the network can capture more abstract and high-level representations of the input video segment.
We evaluate four different types of recurrent layers: long short-term memory (LSTM), gated recurrent units (GRU), bidirectional LSTM (BiLSTM), and bidirectional GRU (BiGRU).

We aim to detect six relevant actions including abdominal access, grasping anatomy, knot pushing, needle pulling, needle pushing, and suction from a large set of gynecologic laparoscopy videos that contain various actions. To train the model for each target action, we use a binary classification approach where the input videos are split into two classes: the target action, and the rest of the actions. We repeat this process for each target action, resulting in a total of six binary classification models. This approach ensures that the model focuses on learning the unique characteristics of the target action and becomes less prone to confusion with other actions existing in the video.

\subsection{Training and Inference Framework} As shown in Fig~\ref{fig:Overview} (bottom), our video dataset consists of raw videos of laparoscopic surgery with their duration being approximately 20 minutes, together with the action labels with a temporal resolution of 25 frames per second. As the first step, we segment the videos into 2-3 second non-overlapping clips using the provided action labels. Afterward, we split the whole dataset into two groups: the target action and the other actions. In this stage, the number of clips corresponding to the target action is usually far smaller than the number of other actions. To address this high imbalance problem, we apply offline augmentations on the videos from the target class to finally achieve the same number of clips in both classes (Data augmentation module). We further adopt a random sampling strategy to provide the maximum diversity in training data from our limited training set. Concretely, we set the length of the network's input sequence to 20 frames. Accordingly, we only require 20 frames from a video clip of length 2-3 seconds (50-75 frames considering a frame rate equal to 25fps). In every iteration, the selected video clips are fed to our frame sampling module. In this module, the input video sequence is split into 20 non-overlapping segments, and one frame is randomly sampled from each segment. This method assures a maximum level of diversity for the generated frame sequences.

%% file: 04_experimental_settings.tex
\begin{table}[t!]
\centering
\caption{Data augmentation techniques employed for the training set}
\label{tbl:label1}
\begin{tabular}{|@{ }c@{ }|@{ }c@{ }|@{ }c@{ }|}
    \specialrule{.12em}{.05em}{.05em}  
         Augmentation technique & Parameter & Value\\
    \specialrule{.12em}{.05em}{.05em}  
         Gamma contrast & Gamma value & (0.5)\\
    
         Gaussian blur & Sigma & (10)\\
    
         Brightness & Range & (0.2, -0.2)\\
    
         Saturation & Saturation value & (1.5)\\
         
         Horizontal Flip & Probability & 0.5\\
    \specialrule{.12em}{.05em}{.05em}
    
\end{tabular}
\end{table}
\begin{table*}[t!]
\renewcommand{\arraystretch}{1}
\caption{Quantitative comparisons of the proposed CNN-RNN models and static CNNs based on accuracy.}
\label{tbl:label2}
\centering
\begin{tabular}{lp{2cm}p{1.5cm}p{1.5cm}p{1.5cm}p{1.5cm}p{1.5cm}p{1.5cm}p{1.5cm}}
\specialrule{.12em}{.05em}{.05em}
Backbone  & Head & Abdominal Access & Grasping Anatomy& Knot Pushing& Needle Pulling& Needle  Pushing& Suction & Average\\\specialrule{.12em}{.05em}{.05em}
\multirow{5}{*}{VGG16} & Fully Connected &52.15 & 46.87 & 49.43 & 46.91 & 48.64 & 50.23 &  49.04\\
 &LSTM &86.24 & 86.31 & 76.39 & 73.40 & 72.41 & 85.25 & 80.00\\
 &GRU &85.32& 88.10 & 70.14 & 72.34 & 75.00 & 81.66 & 78.76\\
 &BiLSTM& 87.16 & 88.69 & 77.08 & 75.53 & 76.72 & 88.76 & 82.32\\
 &BiGRU &88.99& 87.50 & 75.69 & 73.30 & 78.45 & 88.17 & 82.01\\
\specialrule{.12em}{.05em}{0.05em}
\multirow{5}{*}{ResNet50} &Fully Connected &60.44 & 56.06 & 51.20 & 64.99 & 63.05& 51.51 & 57.87\\ 
&LSTM &\textbf{91.74} & 92.86 & 79.86 & 78.72 & 80.17& 80.71 & 84.01\\
 &GRU &88.07& 91.67 & 79.85 & 76.60 & 77.59& 82.86 &  82.77\\
 &BiLSTM&90.83& \textbf{93.45} & \textbf{87.50} & 77.66 & \textbf{81.90} &\textbf{89.35} & \textbf{86.78}\\
 &BiGRU &89.91& 91.07 & 77.80 & \textbf{82.98}& 81.03 & 85.00 &  84.63\\
\specialrule{.12em}{.05em}{0.05em}
\multirow{5}{*}{EfficientNetB2} &Fully Connected &48.24 & 53.57 & 52.11 & 49.30 & 51.02& 49.85 & 50.68\\
 &LSTM &87.16 & 70.83 & 73.61 & 79.79 & 78.45& 70.71 & 76.75\\
 &GRU &83.49& 69.64 & 72.92 & 74.47 & 78.40& 66.43 &  74.22\\
 &BiLSTM&90.83& 73.21 & 77.78 & 76.60 & 84.48 &72.86 & 79.29\\
 &BiGRU &89.90& 66.67 & 72.22 & 81.91 & 81.03 &  65.71 & 76.24\\
\specialrule{.12em}{.05em}{0.05em}
\multirow{5}{*}{DenseNet121} &Fully Connected &57.34 & 46.60 &45.35 & 43.90 & 49.67&  49.64 & 48.75\\
 &LSTM &78.90 & 72.62 &86.81 & 77.66 & 74.14&  79.29 & 78.23\\
 &GRU &75.23& 72.02 & 77.08 & 76.6 & 70.69&  80.71 & 75.38\\
 &BiLSTM&82.57& 76.19 & 89.58 & 79.80 & 81.03 & 83.57 & 82.12\\
 &BiGRU &77.06& 79.17 & 87.50 & 71.27 & 77.59 & 81.43 & 79.00\\
\specialrule{.12em}{.05em}{0.05em}
\end{tabular}
\end{table*}
\section{Experimental Settings}
\label{sec: experimental settings}

\paragraph{Dataset}

Our dataset includes 18 laparoscopic surgery videos randomly sampled from a collection of more than 600 recordings of gynecologic laparoscopy at the Medical University of Vienna with a resolution of $1920\times 1080$, which is down-scaled to $224\times224$. All 18 laparoscopic surgery videos are segmented and annotated by clinical experts as abdominal access, grasping anatomy, knot pushing, needle pulling, needle pushing, suction, and the ``other actions" class. It should be noted that some laparoscopic videos do not contain all different actions.
For the training set per each target action, we use about 30 video clips from the target action and 180 to 200 clips from the rest of the actions. Regarding this, five videos from each category are selected for training and two other videos for testing.

\paragraph{Data Augmentation Methods}
Video augmentation is a powerful technique that can improve the performance of video classification models. It also increases the diversity of training data available to the model, which helps to reduce overfitting and improve the generalization of the model. In our study, we used video augmentation to generate additional training samples by applying transformations such as gamma contrast, gaussian blur, brightness, saturation, and horizontal flip to the videos. A detailed description of the applied  augmentations is provided in Table~\ref{tbl:label1}.

\paragraph{Alternative Methods}

To confirm the effectiveness of our proposed CNN-RNN model, we also train static CNN models on the same dataset. We evaluate the same popular CNN architectures that we use in our proposed network architecture, e.g., VGG16, ResNet50, EfficientNetB2, and DenseNet121, which have shown excellent performance in various computer vision tasks, including image classification. 

\paragraph{Neural Network Settings}
In all settings, the backbone network used for feature extraction is pretrained on the ImageNet dataset.
Regarding the static CNNs,
we add several new layers on top of the backbone network. Specifically, we add an average pooling layer, followed by a dense layer with 256 units, a dropout layer with a rate of 0.5, a dense layer with 64 units, and finally, a two-unit dense layer and softmax activation function. 
For CNN-RNN models, the RNN component is designed with two layers: The first layer with 128 units enables the model to incorporate high-level representations of the features corresponding to independent frames. The second layer with 64 units provides a more refined representation by capturing the subtle temporal dynamics of the video frames. Using two stacked recurrent layers, the RNNs are able to learn more complex temporal patterns from the video data. However, increasing the number of parameters as a result of stacking two recurrent layers increases the risk of overfitting. Hence, we use dropout regularization between the layers to prevent overfitting. 

\paragraph{Training settings}
We train each of the CNN-RNN architectures on our video dataset using the binary cross-entropy loss function and the Adam optimizer with a learning rate of $\alpha = 0.001$. We also use early stopping with patience of 20 epochs to prevent overfitting.

\paragraph{Evaluation Metrics}
To evaluate the performance of our model, we apply two commonly used metrics, namely accuracy and F1-score.

%% file: 05_experimental_results.tex
\section{Experimental Results}
\label{sec: experimental results}
To evaluate the performance of the proposed CNN-RNN model and the static CNNs, we conduct a series of experiments on the laparoscopic surgery videos. In this section, we present the experimental results and analyze the performance of each model in detail. 

Table~\ref{tbl:label2} compares the performance of the proposed CNN-RNN architecture with different CNN networks and recurrent layers, as well as the static CNNs in terms of accuracy. Accordingly, the models using ResNet50 show the highest performance compared to other static CNNs including VGG16, EfficientNetB2, and DenseNet121. This could be attributed to the deeper architecture of ResNet50, allowing it to learn more complex features and patterns in the input frames. In addition, among the different combinations of CNN and RNN architectures that we experiment with, the one that yields the best performance is ResNet50 as the backbone with BiLSTM layers as the RNN. This combination achieves an average accuracy of $86.78\%$, in recognizing different actions in laparoscopic surgery videos, as opposed to $84.63\%$ corresponding to the second-best model (ResNet50-BiGRU). This further confirms the complexity of action recognition task, since BiLSM layers contain more trainable parameters and can memorize longer-term dependencies compared to BiGRU layers. It should be noted that bidirectional layers in all different models have considerably boosted action recognition performance compared to one-directional recurrent layers. To summarize the table, the results suggest that ResNet50 enables the extraction of more robust features from the input frame sequences, while the BiLSTM layer can more effectively model the temporal relationships between the frames.

The bar chart presented in Fig.~\ref{fig:evaluation} provides a comprehensive comparison of the F1-score corresponding to the six target actions for the five Network architectures with the best-performing CNN backbone (ResNet50). The results show that the action 'Grasping anatomy' has achieved the highest F1-score of 93.56\% when compared to the other five actions in the laparoscopic surgery videos. This indicates that the ResNet50-BiLSTM model has performed well in recognizing the grasping anatomy action in the videos. 
Although the Needle Pulling, Needle Pushing, and Knot Pushing actions are known to be very similar and challenging to recognize, the proposed model achieved superior performance in detecting these actions, particularly in the BiGRU model, with the Needle Pulling action obtaining the f1-score of 84.45\%.
Overall, The results from the average bar show that the combination of ResNet50 and stacked BiLSTM layers as the main components outperform all other models in terms of F1-score.

Finally, it should be noted that we cannot compare our achieved performance to other approaches, because no other work so far has focused on such specific short-termed temporal action recognition in gynecologic laparoscopy (see also end of Section~\ref{sec: related work}). However, since we release the dataset with this paper (including all videos and annotations), future works will be able to compare to our results.

%% file: 06_conclusion.tex
\begin{figure}[tbp!]
    \centering
    \includegraphics[width=9cm]{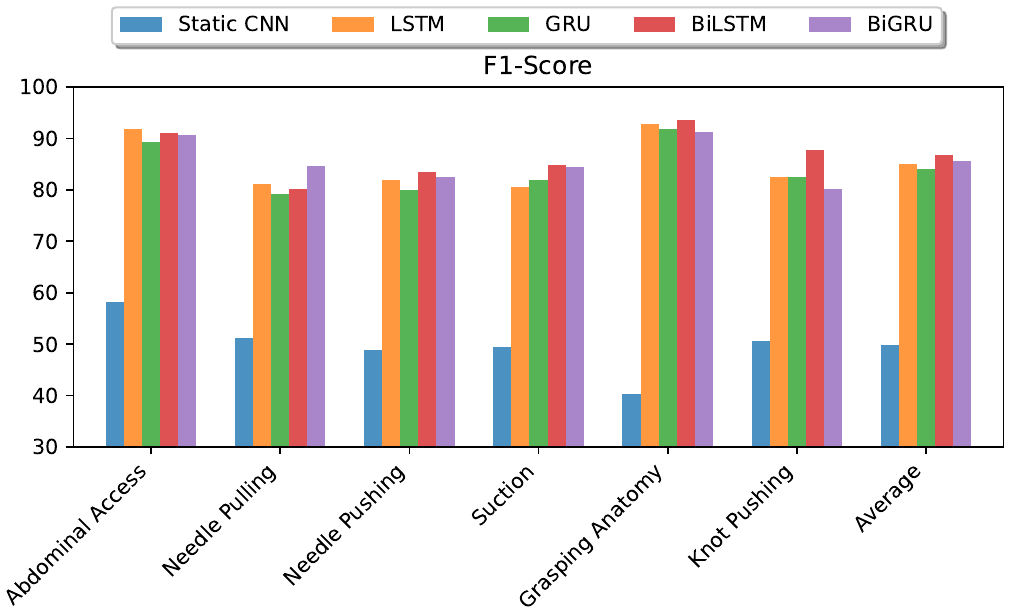}
    \caption{F1-Score corresponding to different architectures with ResNet50.}
    \label{fig:evaluation} 
\end{figure}

\section{Conclusion}
\label{sec: conclusion}

In this study, we presented a neural network architecture and a novel training/inference framework for action classification in gynecologic laparoscopy videos. We have performed comprehensive evaluations using four pre-trained CNN architectures, including VGG16, ResNet50, EfficientNetB2, and DenseNet121, and four types of recurrent layers, including LSTM, GRU, BiLSTM, and BiGRU, to find the best combination for our classification task. To the best of our knowledge, this work presents the first fully automatic approach for action recognition in laparoscopic surgery videos, considering real-world challenges rather than laboratory conditions. Our results reveal that the combination of a deeper CNN architecture, such as ResNet50 backbone with BiLSTM layers, results in the best performance in the mentioned task. Overall, our findings suggest that the proposed model can be an effective tool for video classification in laparoscopic surgery videos, enabling fully-automatic segmentation of these videos for follow-up usages such as surgical training, surgical skill evaluation, outcome estimation, etc. Future work will focus on extending the study to improve cross-dataset action recognition performance and enhance the generalization performance in low-data regimes.